%% file: main.tex
\documentclass[letterpaper, 10 pt, conference]{ieeeconf}

\IEEEoverridecommandlockouts
\usepackage{cite}
\usepackage{amsmath,amssymb,amsfonts}
\usepackage{algorithmic}
\usepackage{graphicx}
\usepackage{textcomp}
\usepackage{xcolor}
\usepackage{booktabs}
\usepackage{tabularx}
\usepackage{makecell}
\usepackage{subcaption}
\usepackage[font=small]{caption}
\usepackage{multirow}
\usepackage{pifont}
\usepackage[table]{xcolor}
\usepackage[hidelinks]{hyperref}
\newcommand{\cmark}{\ding{51}}
\newcommand{\xmark}{\ding{55}}

\definecolor{stageBlue}{RGB}{79,129,189}   % clearer blue
\definecolor{stageRed}{RGB}{192,80,77}     % clearer red
\definecolor{stageGreen}{RGB}{155,187,89}  % clearer green

\def\BibTeX{{\rm B\kern-.05em{\sc i\kern-.025em b}\kern-.08em
    T\kern-.1667em\lower.7ex\hbox{E}\kern-.125emX}}
\begin{document}

\title{\LARGE \bf
Learning Tactile-Aware Quadrupedal Loco-Manipulation Policies
}

\author{
Pokuang Zhou$^{1}$, 
Yuhao Zhou$^{1}$, 
Quan Khanh Luu$^{1}$, 
Seungho Han$^{1}$, 
Heng Zhang$^{1,2}$, \\
Binghao Huang$^{3}$, 
Yunzhu Li$^{3}$, 
Arash Ajoudani$^{2}$, 
Zhengtong Xu$^{1,\dagger}$, 
and Yu She$^{1,\dagger}$%
\thanks{$^{1}$Purdue University, West Lafayette, IN, USA. $^{2}$Istituto Italiano di Tecnologia, Genoa, Italy. $^{3}$Columbia University, New York, NY, USA. $^\dagger$Equal advising. This work was supported by the National Science Foundation under Grants 2423068 and 2520136, and the United States Department of Agriculture under Grants 2023-67021-39072 and 2024-67021-42878. This work used GPU resource from NSF ACCESS CIS 260072.}%
}

\maketitle

\begin{abstract}
Quadrupedal loco-manipulation is commonly built on visual perception and proprioception. Yet reliable contact-rich manipulation remains difficult: vision and proprioception alone cannot resolve uncertain, evolving interactions with the environment. Tactile sensing offers direct contact observability, but scalable tactile-aware learning framework for quadrupedal loco-manipulation is still underexplored.
In this paper, we present a tactile-aware loco-manipulation policy learning pipeline with a hierarchical structure. Our approach has two key components. First, we leverage real-world human demonstrations to train a tactile-conditioned visuotactile high-level policy. This policy predicts not only end-effector trajectories for manipulation, but also the evolving tactile interaction cues that characterize how contact should develop over time. Second, we perform large-scale reinforcement learning in simulation to learn a tactile-aware whole-body control policy that tracks diverse commanded trajectories and tactile interaction cues, and transfers zero-shot to the real world. Together, these components enable coordinated locomotion and manipulation under contact-rich scenarios.
We evaluate the system on real-world contact-rich tasks, including in-hand reorientation with insertion, valve tightening, and delicate object manipulation. Compared to  vision-only and visuotactile baselines, our method improves performance by 28.54\% on average across these tasks. The project website is available at
\href{https://pokuangzhou.github.io/tactile-aware-quadrupedal-loco-manipulation/}{\nolinkurl{https://pokuangzhou.github.io/tactile-aware-quadrupedal-loco-manipulation/}}.
\end{abstract}

%%%%%%%%%%%%%%%%%%%%%%%    sections   %%%%%%%%%%%%%%%%%%%%%%%%%%%%%%%%%%%%%%%%%%%%%

\input{sections/1introduction.tex}

\input{sections/2relatedwork.tex}

\input{sections/3method.tex}

\input{sections/4experiment.tex}

\input{sections/5conclusion.tex}

%%%%%%%%%%%%%%%%%%%%%%%%%%%%%%%%%%%%%%%%%%%%%%%%%%%%%%%%%%%%%%%%%%%%%%%%%%%%%%%%%%%

\bibliographystyle{IEEEtran}
\bibliography{reference}

\end{document}

%% file: sections/1introduction.tex
\section{INTRODUCTION}

Loco-manipulation based on quadrupedal mobile manipulators substantially expands a robot’s operational workspace and task repertoire, making these platforms promising for real-world deployment in inspection, maintenance, and field robotics applications~\cite{fu2023deep, gong2023legged,zhang2026SurveyPhysicalAI}. Recent quadrupedal loco-manipulation systems are commonly built on vision and proprioception, learning coordinated base–arm behaviors from data. However, reliable and generalizable contact-rich manipulation remains challenging: vision and proprioception lack direct access to the rich, precise contact information needed to infer evolving interaction dynamics at the interface, and vision is further degraded by occlusion and partial observability.

A growing body of evidence shows that tactile feedback substantially improves precision and robustness during physical interaction~\cite{zhang2026touchguide, xu2025unit, hu2023machine}. However, making quadrupedal loco-manipulation tactile-aware at scale remains challenging: tactile signals are high-dimensional and strongly coupled with whole-body dynamics, and contact-rich tasks require reasoning over how interaction evolves over time while the robot simultaneously maintains stable locomotion. This creates a mismatch between where tactile information is most valuable (contact evolution and manipulation intent) and where it must be realized (whole-body, contact-constrained execution), making it difficult to learn policies that are both reliable and transferable.

In this paper, we propose a tactile-aware hierarchical loco-manipulation learning framework that integrates tactile sensing into both high-level planning and low-level whole-body control. Our primary contributions are:

{\setlength{\textfloatsep}{6pt}
\begin{figure}[t]
  \centering
  \includegraphics[width=0.49\textwidth]
  {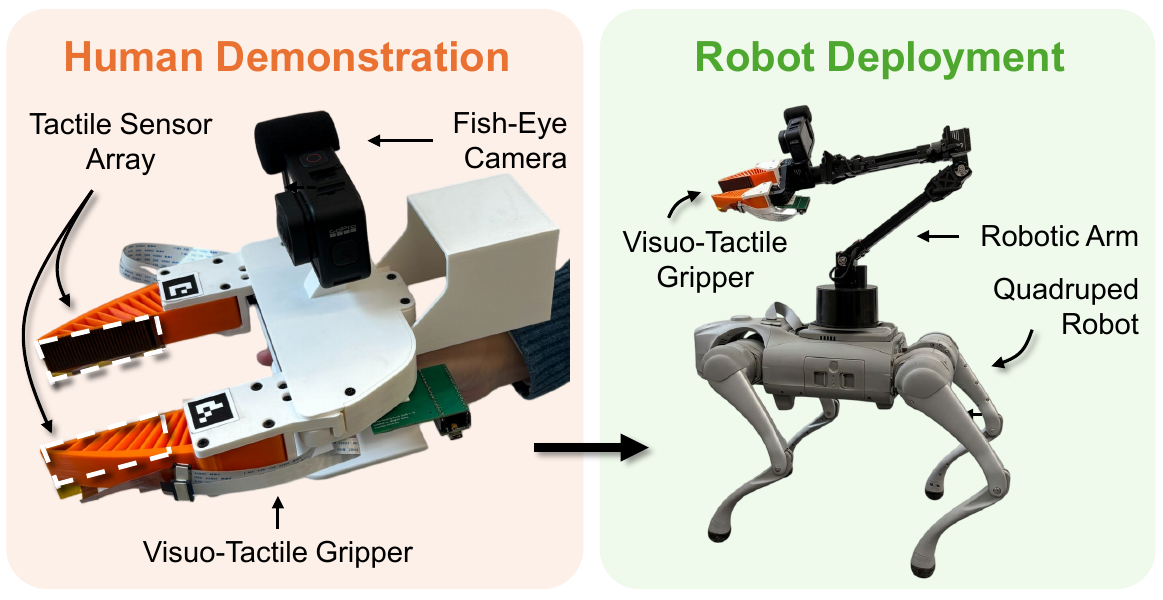}
  \caption{We achieve fully autonomous real-world tactile-aware quadrupedal loco-manipulation by learning from demonstrations collected with a hand-held, tactile-instrumented UMI gripper. Our approach enables learning and deployment across tasks and objects, supporting contact-rich manipulation (e.g., in-hand reorientation with insertion and valve tightening) as well as damage-free handling of delicate items such as fragile chips and fruits.}
  \label{fig:main_fig}
\end{figure}
}

1. \textbf{A deployable, end-to-end visuotactile framework for quadrupedal loco-manipulation}.
We design a hierarchical learning architecture that leverages both vision and tactile sensing to enable real-world, contact-rich loco-manipulation on quadrupedal mobile manipulators. We evaluate the system on diverse contact-intensive tasks, including in-hand marker reorientation with subsequent insertion, valve tightening, and delicate object manipulation, and achieve a 28.54\% average performance improvement over visuomotor and visuotactile baselines.

2. \textbf{A tactile-aware hierarchical policy learning that bridges contact evolution and whole-body execution.}
We propose a tactile-aware hierarchical loco-manipulation framework that integrates tactile sensing into both the manipulation policy and the whole-body control policy. At the high level, we train a tactile-conditioned diffusion model that predicts not only end-effector pose trajectories but also future tactile interaction cues, explicitly modeling the evolution of contact. At the low level, we use large-scale reinforcement learning in simulation to train a tactile-aware whole-body policy that tracks these commands under contact constraints while maintaining stable and efficient locomotion, and transfers zero-shot to the real world.

This paper is structured as follows. Section~\ref{sec:related works} reviews related works. Section~\ref{sec: method} presents our method, including a high-level and a low-level policy. Section~\ref{sec: experiments} reports real-world experimental results and analyzes. Section~\ref{sec: conclusion} concludes the paper and outlines future directions.

\begin{figure*}[!t]
  \centering
  \includegraphics[width=0.98\textwidth]
  {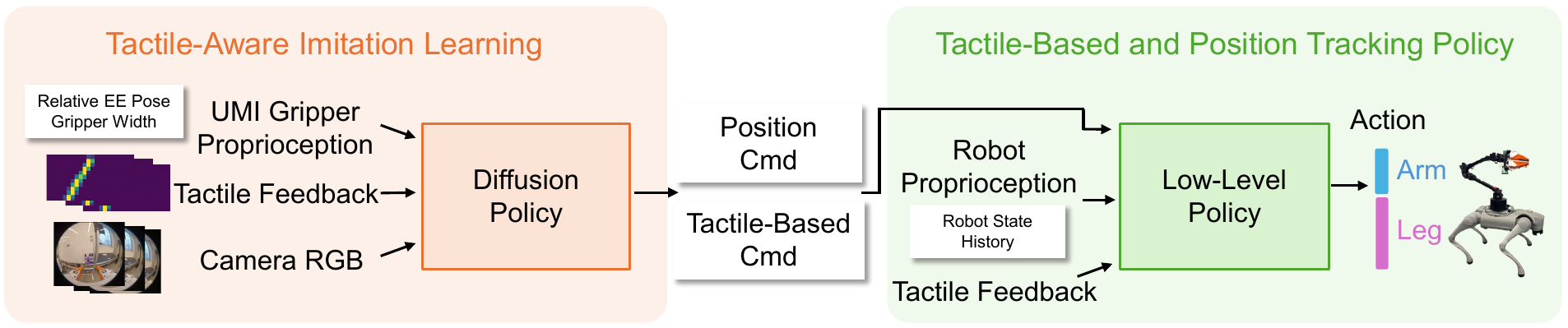}
  \caption{Pipeline of our proposed visuotactile loco-manipulation policy learning framework. The high-level policy, trained from human demonstrations via imitation learning, predicts position and tactile commands. The low-level controller, trained in simulation with RL, takes the robot state, tactile feedback, and action history as input to track the position and tactile commands generated by the high-level policy. EE denotes end-effector.}
  \label{fig:pipeline}
\end{figure*}

%% file: sections/2relatedwork.tex
\section{Related Works}\label{sec:related works}

To enable low-cost and portable demonstration collection in diverse real-world settings, recent work has adopted UMI-style handheld gripper interfaces \cite{chi2024universal}. This paradigm supports in-the-wild data collection and enables hardware-agnostic diffusion policies that transfer across robot embodiments. Building on the same interface, \cite{zhu2025touch} further incorporates tactile arrays into the gripper to capture vision and touch simultaneously.

A growing body of work has shown that tactile feedback is particularly valuable for contact-rich manipulation, supporting tasks such as power-plug insertion \cite{yu2023mimictouch,xie2025towards,fu2023safe}, light-bulb screwing \cite{luu2025manifeel}, cable untangling \cite{zhou2025inhand}, and peeling \cite{zhang2026touchguide}. For quadrupedal robots, \cite{lin2025locotouch} presents a system that mounts tactile sensors on the robot’s back to stabilize a cylindrical object during balancing. 

By combining the mobility of the quadruped base with the dexterity of the manipulator, whole-body control (WBC) enables integrated loco-manipulation with substantially expanded reach and task versatility \cite{ha2024umi, margolis2023walk, zhi2025learning, portela2025whole}. With the growing use of end-to-end reinforcement learning (RL), learning-based low-level policies for tracking dense reference motions have become increasingly prevalent. Early work introduced an RL-based regularized online adaptation framework and demonstrated successful sim-to-real transfer for quadruped loco-manipulation \cite{fu2023deep}. Building on this line of research, later studies further improved manipulation accuracy and overall performance through optimized command sampling \cite{portela2025whole}, arm inverse kinematics (IK) for more precise end-effector control \cite{liu2024visual}, and two-stage policy architectures that decouple locomotion and manipulation objectives \cite{10925884}.

More recently, hierarchical RL frameworks have been introduced to incorporate active perception into high-level decision making. These methods integrate visual perception either through jointly trained RL policies or separate perception models \cite{liu2024visual, wang2025quadwbg}. In parallel, semantically grounded large language model-based planners have been explored for task-level reasoning and command generation \cite{11155187, wang2025odyssey}. Human-demonstration-based imitation learning, especially diffusion policy methods, has also been used to produce high-level commands that guide the underlying whole-body controller \cite{liu2025mlm, ha2024umilegs}.

However, these methods remain limited for contact-rich manipulation, where successful execution requires iterative adjustment of an object’s pose while maintaining stable contact. In such settings, vision alone is often insufficient to capture the fine-grained contact state needed for reliable control~\cite{luu2025manifeel}. To address this limitation, we propose a tactile-aware policy learning framework for quadrupedal loco-manipulation.

%% file: sections/3method.tex
\section{Method}
\label{sec: method}

In this section, we introduce the overall framework, including the high-level policy and the tactile-aware low-level policy, to enable visuotactile contact-rich loco-manipulation. The whole pipeline of is shown in Fig~\ref{fig:pipeline}.

\subsection{Tactile-Aware Diffusion Policy}

We collect teleoperated demonstrations using a UMI gripper~\cite{zhu2025touch}.
We represent the UMI gripper state as a 6-DoF end-effector pose expressed in the robot base (arm-base) frame.
At each time step $t$, the policy observes an on-wrist RGB image, UMI gripper pose, and low-dimensional tactile features.
Concretely, we define the observation as $o_t=(i_t,x_t,w_t,s_t)$, where $i_t$ is the RGB image, $x_t$  is the UMI gripper pose, $w_t$ is the gripper width, and $s_t \;=\; (m_t,\; \theta_t,\; c_t)$  is a low-dimensional tactile descriptor consisting of contact area $m_t$, contact orientation $\theta_t$, and contact center $c_t \in \mathbb{R}^2$
(in the tactile image or gripper frame). The tactile signals are obtained from the $12\times 32$ tactile sensors~\cite{zhu2025touch}.

We encode the RGB image $i_t$ with a CLIP-pretrained ViT that is fine-tuned on our task~\cite{zhu2025touch}, and concatenate it with the normalized low-dimensional signals $(x_t, w_t, s_t)$ to form a compact observation.

The resulting visual embedding is concatenated with the normalized signals to form the conditioning input to the diffusion policy \cite{chi2025diffusion}.
Following common practice, the policy is conditioned on a short history of the past
% $o$ 
observations to capture immediate temporal context.

% \subsubsection{Action Representation.}

Rather than predicting only the UMI gripper pose, the diffusion policy predicts a short-horizon sequence of low-dimensional targets that includes both end-effector motion commands and tactile targets over a horizon of length $H$, denoted as $a_{t:t+H-1}=({a_t,a_{t+1},\ldots,a_{t+H-1}})$, where each action is defined as $a_t=(x^{d}_{t+1},w^{d}_{t+1},s^{d}_{t+1})$. Here, $x^{d}_{t+1}$ is the target end-effector pose for the next control step, $w^{d}_{t+1}$ is the target gripper width, and $s^{d}_{t+1}=(m^{d}_{t+1},\theta^{d}_{t+1},c^{d}_{t+1})$ denotes the desired tactile target.

For the diffusion model, the denoiser is parameterized with a U-Net backbone, with conditioning injected via FiLM~\cite{perez2018film}. We use DDIM~\cite{song2020denoising} for inference and perform rollout with receding-horizon control~\cite{chi2025diffusion}. The predicted pose and tactile targets are then tracked by the proposed low-level policy.

\subsection{Tactile-Aware Low-level Policy}
This section describes how we design the low-level policy to execute the position- and tactile-based commands produced by the high-level policy.
We use a two-stage curriculum: train stable locomotion and command tracking in simplified environments, then train in task-specific scenes that match the real setup for efficient sampling and robust contact-rich execution.

{\setlength{\textfloatsep}{4pt}\setlength{\intextsep}{6pt}
\begin{figure}[!t]
  \centering
  \includegraphics[width=1\columnwidth]
  {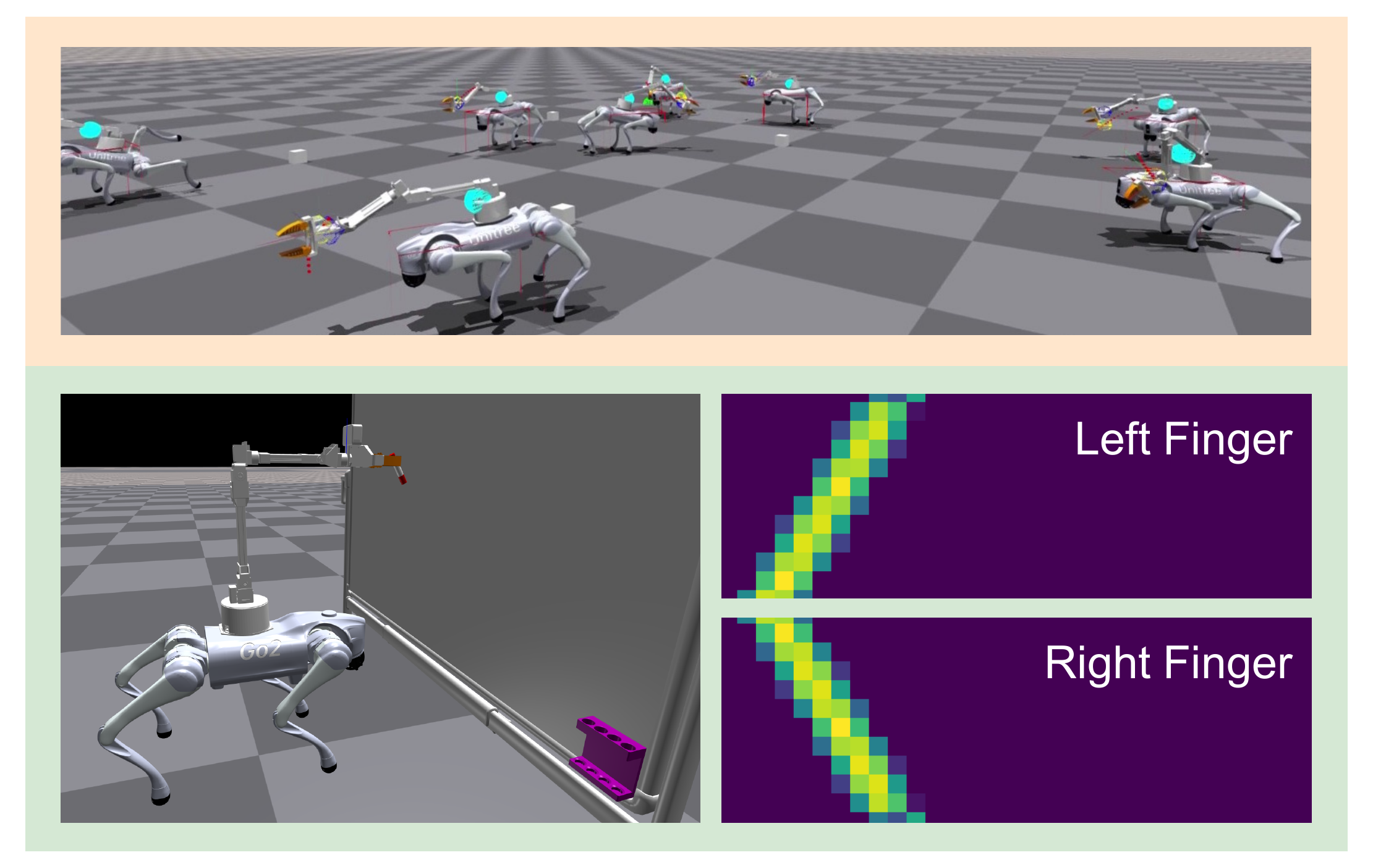}
  \caption{Low-level policies are trained in simulation. Top: stable locomotion with pose-tracking training. Bottom: tactile-aware loco-manipulation policy training built on top of the pose-tracking policy. }
  \label{fig:simulation}
\end{figure}
}

\subsubsection{Stable Locomotion Policy}

In this stage, we train a low-level policy for stable locomotion while tracking randomly sampled trajectories. The quadruped is trained to track randomly sampled base velocity commands. We generate end-effector targets by sampling waypoint sequences in spherical coordinates centered at the arm base. Following the low-level goal-reaching formulation in Visual Whole-Body Control (VBC)~\cite{liu2024visual}, the policy is conditioned on base-motion commands together with an end-effector pose target, and observes a compact proprioceptive state (base/arm/leg kinematics, previous action, and phase/timing signals) for robust tracking.
For precise arm control, we realize the end-effector waypoints via IK, converting each waypoint into joint-position targets.

We train the low-level policy in Isaac Gym using massively parallel simulated environments, with a Trimesh terrain curriculum and extensive domain randomization (as detailed in the Table~\ref{tab:domain_randomization_ranges}).
 All command randomization is defined in the quadruped base frame. In addition, we inject external perturbations by periodically applying random planar velocity pushes to the base, with a maximum magnitude of $0.5$~m/s every $3$~s. 

\begin{table}[t]
\centering
\footnotesize
\caption{Domain randomization ranges used in simulation.}
\setlength{\tabcolsep}{6pt}
\begin{tabular}{l l}
\toprule
\textbf{Randomization item} & \textbf{Range} \\
\midrule
Ground friction coefficient & $[0.3,\,3.0]$ \\
Base mass offset (kg) & $[-0.5,\,2.5]$ \\
Base CoM shift per axis (m) & $[-0.15,\,0.15]$ \\
Motor-strength scaling (legs \& arm) & $[0.7,\,1.3]$ \\
Gripper mass offset (kg) & $[0,\,0.3]$ \\
Linear velocity command $v_x, v_y$ (m/s) & $[-1.5,\,1.5]$ \\
Yaw rate command $\omega_z$ (rad/s) & $[-1.0,\,1.0]$ \\
Initial base position perturbation (m) & within $0.5$ \\
Initial yaw perturbation (rad) & $\pm \pi/2$ \\
Initial base velocity perturbation (m/s) & within $0.1$ \\
Initial joint-angle scaling & $[0.8,\,1.2]$ \\
\bottomrule
\end{tabular}
\label{tab:domain_randomization_ranges}
\end{table}

These randomizations are applied to reduce the sim-to-real gap and improve locomotion robustness. We adopt a reward design following prior work \cite{fu2023deep,ha2024umi,liu2024visual} and tune the weights (Table~\ref{tab:symbols_low_level}, Table~\ref{tab:reward_loco_body}). For brevity, we omit some stability/safety, regularization and regularization terms, all other settings primarily follow Walk These Ways \cite{margolis2023walk}, as well as VBC \cite{liu2024visual} and UMI on Legs \cite{ha2024umi}. 
This stage is critical because arm motion continuously shifts the robot’s center of mass while tracking commanded linear and angular velocities, requiring robust balance and stable whole-body coordination.

\begin{table*}[t]
\centering
\scriptsize

% ---------- Left table ----------
\begin{minipage}[t]{0.34\textwidth}
\centering
\captionof{table}{Definition of symbols.}
\label{tab:symbols_low_level}
\begin{tabularx}{\linewidth}{@{}>{\raggedright\arraybackslash}X l@{}}
\toprule
\textbf{Name} & \textbf{Symbol} \\
\midrule
Leg joint positions          & $\mathbf{q}_{\text{leg}}$ \\
Default joint positions      & $\mathbf{q}_{\text{default}}$ \\
Leg joint velocities         & $\dot{\mathbf{q}}_{\text{leg}}$ \\
Joint torques                & $\boldsymbol{\tau}$ \\
Base linear velocity         & $\mathbf{v}_{b}$ \\
Base angular velocity        & $\boldsymbol{\omega}_{b}$ \\
Base linear velocity command & $v^{*}_{x}$ \\
Base yaw velocity command    & $\omega^{*}_{\text{yaw}}$ \\
Feet contact force           & $\mathbf{f}_{\text{foot}}$ \\
Feet velocity                & $\mathbf{v}_{\text{foot}}$ \\
Feet air time                & $t_{\text{air}}$ \\
Desired contact state        & $c^{\text{cmd}}_{\text{foot}}$ \\
Gait phase / timing offset   & $\boldsymbol{\theta}^{\text{cmd}}$ \\
Default hung-up leg joint positions & $\mathbf{q}^{\text{hung}}_{\text{leg}}$ \\
\bottomrule
\end{tabularx}
\end{minipage}
\hfill
% ---------- Right table ----------
\begin{minipage}[t]{0.65\textwidth}
\centering
\captionof{table}{Key reward functions.}
\label{tab:reward_loco_body}
\begin{tabularx}{\linewidth}{@{}>{\raggedright\arraybackslash}p{0.31\linewidth}
>{\raggedright\arraybackslash}X
r@{}}
\toprule
\textbf{Name} & \textbf{Definition} & \textbf{Weight} \\
\midrule
\multicolumn{3}{@{}l}{\textit{Command-Following Rewards}} \\
\midrule
Linear velocity tracking (x) &
  $\varphi\!\left(v^{*}_{x} - v_{b,x}\right)$ &
  $2.0$ \\
Yaw velocity tracking &
  $\varphi\!\left(\omega^{*}_{\text{yaw}} - \omega_{b,\text{yaw}}\right)$ &
  $0.5$ \\
\midrule
\multicolumn{3}{@{}l}{\textit{Gait \& Contact Rewards}} \\
\midrule
Feet air time &
  $\sum_{i=1}^{4}(t_{\text{air},i} - 0.5)\mathbf{1}[\text{first contact}]$ &
  $2.0$ \\
Feet height &
  $\operatorname{clamp}(\|\mathbf{h}_{\text{foot}}\| - h^{\text{target}}_{\text{foot}}, 0)$ &
  $1.0$ \\
Swing phase tracking (force) &
  $\sum_{\text{foot}}(1-c^{\text{cmd}}_{\text{foot}})
  \left(1-\exp(-\|\mathbf{f}_{\text{foot}}\|^{2}/\sigma_{cf})\right)$ &
  $-2.0$ \\
Stance phase tracking (velocity) &
  $\sum_{\text{foot}}c^{\text{cmd}}_{\text{foot}}
  \left(1-\exp(-\|\mathbf{v}_{\text{foot}}\|^{2}/\sigma_{cv})\right)$ &
  $-2.0$ \\
\midrule
\multicolumn{3}{@{}l}{\textit{Regularization Rewards}} \\
\midrule
Joint torques &
  $\|\boldsymbol{\tau}\|^{2}$ &
  $-2.5{\times}10^{-5}$ \\
Joint acceleration &
  $\|(\dot{\mathbf{q}}_{t}-\dot{\mathbf{q}}_{t-1})/\Delta t\|^{2}$ &
  $-7.5{\times}10^{-7}$ \\
Action rate &
  $\|\mathbf{a}_{t}-\mathbf{a}_{t-1}\|^{2}_{\text{leg}}$ &
  $-0.015$ \\
Feet jerk &
  $\sum_{i}\|\mathbf{f}_{\text{foot},i}-\mathbf{f}_{\text{foot},i}^{\text{prev}}\|$ &
  $-2.0{\times}10^{-4}$ \\
Feet drag &
  $\sum_{i}c_{\text{contact},i}\|\mathbf{v}_{\text{foot},i}\|_{1}$ &
  $-0.08$ \\
\bottomrule
\end{tabularx}
\end{minipage}

\end{table*}

\subsubsection{Tactile-Aware Loco-Manipulation Policy}

In the second stage, we train a tactile-aware low-level policy on top of the first-stage policy. Simulation scenes are constructed to closely match the real setup (as shown in Fig.~\ref{fig:simulation}), such that contacts between the quadruped and the object generate physically realistic interactions.

For trajectory sampling, spherical-coordinate trajectories centered at the arm base are used in the previous stage. This simplified setting allows the policy to learn coordinated quadruped–arm control based on position tracking.
In this stage, we focus on precise tracking of both position- and tactile-based commands. To better match the relative UMI gripper pose commands and improve sampling efficiency, we switch to Cartesian coordinates for trajectory representation. 
We sample UMI gripper trajectories from demonstration data and use a heuristic reachability function to generate quadruped base velocity commands. For a sequence of future end-effector targets, the function evaluates whether the predicted trajectory extends beyond a conservative boundary of the current arm workspace. If so, the exceedance errors of the future targets are accumulated and linearly mapped to the quadruped base motion, producing forward and lateral velocity commands. The generated linear velocities are bounded within 0.3--0.6~m/s.
Based on the sampling method in~\cite{portela2025whole}, we define the arm workspace accordingly.
Considering the need for the end-effector to maintain a suitable orientation, we further reduce the workspace to 70\% of its original size and define this boundary as $d_{\mathrm{thresh}}$, which represents the reachability boundary of the arm workspace.
For the vertical direction, by adjusting the weights of the existing reward terms (see Table~\ref{tab:reward_loco_body}), the robot learns to adopt a tilted posture to enlarge the reachable workspace of the arm.
% In Table~\ref{tab:symbols_low_level} and Table~\ref{tab:reward_loco_body}, we adopt the reward design from VBC~\cite{liu2024visual} and UMI on Legs~\cite{ha2024umi}, and omit overlapping terms for brevity. 

Considering fine-manipulation tasks, we additionally include a balance reward to constrain the quadruped's base motion once the arm starts fine manipulation, which improves the stability of the arm base (see the additional reward terms in Table~\ref{tab:assistant rewards}). Overall, this setting is different from the previously described case where forward and lateral base velocities are generated. Here, when the forward and lateral velocities are set to zero, the quadruped does not translate but instead adjusts its posture, such as the base inclination, to better support fine arm manipulation while remaining as stable as possible. This is important because, during contact-rich fine manipulation, large quadruped base motions can negatively affect manipulation performance.
The reward terms for tracking relative position commands and tactile-based commands are summarized in Table~\ref{tab:assistant rewards}, where $\sigma_{\text{tac}}$ and $\sigma_{\text{pos}}$ denote the tolerance parameters in the exponential tracking rewards. The corresponding reward weights are tuned to balance tracking performance and stability.

% For insertion task, the robot may need multiple attempts to adjust around the hole entrance, which induces rapid changes in tactile signals. For valve tightening, stable tactile feedback requires the UMI gripper to grasp the valve approximately perpendicular to its surface; at the final tightening moment, we observe clear tactile signatures (e.g., a smaller contact area on one finger and a slightly larger contact area on the other). 

% For trajectory sampling, we use spherical-coordinate trajectories centered at the arm base in previous stage. This simplifies early UMI gripper tracking and helps the policy learn coordinated quadruped–arm control. In this stage, we focus on precise tracking of both position- and tactile-based commands, where spherical coordinates become inconvenient; thus, we switch to Cartesian coordinates for trajectory representation.
% We filter the sampled motion commands (i.e., randomized base-velocity and end-effector trajectory targets introduced in the previous section) and only retain trajectories that are similar to those observed in human demonstrations.

% For trajectory sampling, we use spherical-coordinate trajectories centered at the arm base in previous stage. This simplifies early UMI gripper tracking and helps the policy learn coordinated quadruped–arm control. In this stage, we focus on precise tracking of both position- and tactile-based commands, where spherical coordinates become inconvenient; thus, we switch to Cartesian coordinates for trajectory representation.

% --------------------
\begin{table}[t]
\centering
\footnotesize
\caption{{Details of assistance rewards.}
$s^{\text{cmd}}\in \mathbb{R}^{4}$ and $s\in \mathbb{R}^{4}$
are the commanded and measured tactile signals, respectively.}
\label{tab:assistant rewards}
\begin{tabular}{llr}
\toprule
\textbf{Name} & \textbf{Definition} & \textbf{Weight} \\
\midrule
\multicolumn{3}{l}{\textit{Balance Rewards}} \\
\midrule
% $r_{\text{posture}}$ &
%   $\mathbf{1}\!\left[\|\mathbf{p}^{\text{cmd}}\| > d_{\text{thresh}}\right] \cdot \exp\!\left(-\left|h_{b} - h^{\text{lean}}\right|\right)$ &
%   $0.001$ \\
$r_{\text{base-stable}}$ &
  $-\left(0.2\|\mathbf{v}_{b,xy}\|^{2} + 0.5\|\boldsymbol{\omega}_{b}\|^{2}\right)$ &
  $0.02$ \\
$r_{\text{leg-posture}}$ &
  $-\left(\|\mathbf{q}_{\text{leg}} - \mathbf{q}^{\text{hung}}_{\text{leg}}\|^{2} + 0.1\|\Delta \mathbf{q}_{\text{leg}}\|^{2}\right)$ &
  $0.005$ \\
$r_{\text{smooth}}$ &
  $-\left(\|\Delta \mathbf{v}_{b,xy}\|^{2} + \|\Delta \boldsymbol{\omega}_{b}\|^{2}\right)$ &
  $0.1$ \\
\midrule
\multicolumn{3}{l}{\textit{Track Rewards}} \\
\midrule
$r_{\text{tac}}$ &
  $\exp\!\left(-\|s^{\text{cmd}} - s\|^{2} / \sigma_{\text{tac}}\right)$ &
  $0.1$ \\
$r_{\text{pos}}$ &
  $ \exp\!\left(-\|\mathbf{p}^{\text{cmd}} - \mathbf{p}\|^{2} / \sigma_{\text{pos}}\right)$ &
  $0.06$ \\
\bottomrule
\end{tabular}
\end{table}

%% file: sections/4experiment.tex
\section{Experiments and Results}
\label{sec: experiments}

In this section, we introduce a set of experiments for quantitative analysis to validate the effectiveness of our method. First, we present the overall system setup and implementation details. We evaluate the effectiveness and motivation of our method through three components: task design, baseline design, and experimental evaluation.
Finally, we compare against baseline approaches and discuss why our method achieves better performance on these tasks.

% \quan{this section needs to reorganize: ...
% \begin{itemize}
%     \item Start with some Questions, what do we want to evaluate in this section. Then in the results, try to address those questions in subsections, respectively.
%     \item Should have: A. Experimental setups, B. Results, then B.1, B.2, B.3 to address the questions. 
%     \item In A. Experimental setups: 
%     \begin{itemize}
%         \item A.1: Introduce the tasksuite for benchmarking or ablation studies. State it concisely, use headings to highlight the tasks
%         \item A.2: Introduce the setup for baselines. Here it better to find some descriptive names for Baseline 2 and Baseline 3. Table IIIa looks good, but it'd still better to have a name
%     \end{itemize}
%     \item  In B. Results
%     \begin{itemize}
%         \item B.1 could be devoted to highlight and discuss the results in Table III.b
%         \item B.2 could be for the results in the current Capability demonstrations on a mobile quadruped platform 
%     \end{itemize}
% \end{itemize}
% }

\subsection{Experimental Setups}

\subsubsection{System Setup and Implementation Details} The robot system is shown in Fig.~\ref{fig:main_fig}, which consists of a 12-DOF Unitree Go2 quadruped robot and a 6-DOF D1 robot arm, both powered by the battery of Go2. We customize the D1 arm with UMI grippers  with tactile array sensor and a GoPro camera. We deploy FlexiTac tactile sensor array~\cite{zhu2025touch,huang2025vt,akinola2025tacsl} with a spatial resolution of $32 \times 12$ taxels over a sensing area of $66~\mathrm{mm} \times 25~\mathrm{mm}$. It operates by converting pressure-induced changes in electrical resistance into measurable electrical signals. This corresponds to an effective resolution of approximately $2 \times 2 \mathrm{mm}^2$ per taxel. Our policy runs on a separate desktop’s RTX~4070~Ti via a direct cable connection. 

High-level policies are trained on NVIDIA A100 GPUs. Task~1 contains two atomic subtasks, reorientation and insertion, for which we collect demonstrations and train/evaluate separate policies. Similar to Task~2, each subtask policy is trained for 300 epochs using 100 demonstrations. For the full Task~1 long-horizon sequence, we collect 200 demonstrations and train for 400 epochs. And we set the planning horizon to $H{=}16$ and run the diffusion policy at $2~\mathrm{Hz}$.

Low-level policies are trained on a workstation equipped with an RTX~4070~Ti GPU and a Ryzen~9~7900 CPU; each training run takes approximately 40 hours. 

Since the Unitree D1 arm only accepts joint-position commands at 10~Hz, the low-level policy runs at 200~Hz and outputs joint targets for both the quadruped and the arm. During real-world deployment, we downsample the outputs by taking every 4th step for the quadruped (50~Hz) and every 20th step for the arm (10~Hz), enabling synchronized execution with a shared representation that couples motion targets with tactile outcomes.

\subsubsection{Task Suites}
In this section, inspired by prior studies showing that vision-only policy is often insufficient for some contact-rich manipulation \cite{dong2021tactile, zhang2025safe, ye2026visual,luu2025manifeel,xu2025unit}, and to support a comprehensive evaluation, we introduce a task suite as following:

\textbf{Task 1: Extrinsic-Contact-Based Reorientation and Insertion.} For the first task set, we consider a long-horizon manipulation scenario consisting of three stages. In the initial stage, the quadruped robot grasps a marker and performs in-hand reorientation via extrinsic contacts to adjust the marker’s orientation for a subsequent insertion. This reorientation is necessary because an improper orientation prevents successful insertion in a confined workspace~\cite{zhao2025tactile, bronars2024texterity}. In our setup, a vertical whiteboard limits the arm’s forward reach, and the socket’s tight, vertically aligned hole requires in-hand rotation. In next stage, the robot walks to the socket location. In final stage, since the socket height is low and outside the reachable workspace in the hung-up posture, the quadruped performs a lean-down motion to extend its operational workspace. and then execute the insertion. After completing the task, the robot returns to the hung-up posture, and exits the task area.

\textbf{Task 2: Valve Tightening.} For the quadrupedal valve-tightening task, we design a set of challenging scenarios. Starting from the hung-up posture, the valve is initially outside the end-effector workspace for direct manipulation. The quadruped therefore needs to execute a lean-down motion to bring the EE into a feasible operating region, after which the end-effector grasps the valve and performs tightening. During tightening, it is critical to maintain an approximately perpendicular contact between the gripper and the valve. This contact configuration yields clearer and more informative tactile feedback. The tightening procedure follows a cyclic pattern of grasp--twist--release, repeated until the valve reaches the fully tightened state. With tactile feedback enabled, we observe an asymmetric two-finger contact pattern during tightening: one finger exhibits a larger contact area than the other. We attribute this to gripper compliance, which leads to uneven normal-force distribution across the two tactile pads under near-perpendicular contact.

{\setlength{\textfloatsep}{4pt}\setlength{\intextsep}{6pt}
\begin{figure}[t]
  \centering
  \includegraphics[width=0.98\columnwidth]
  {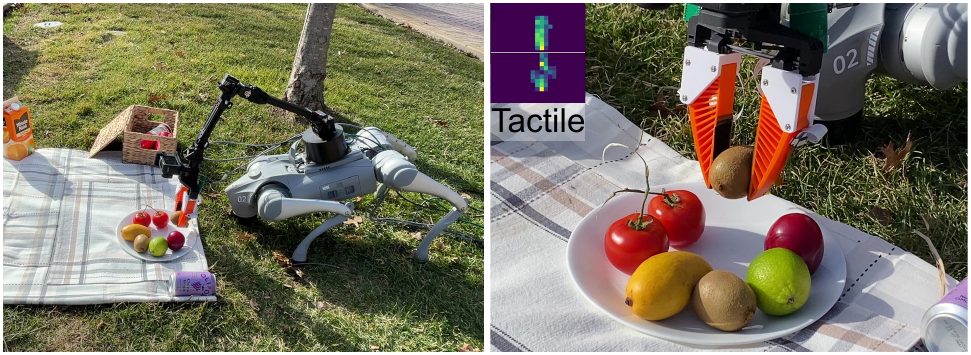}
  \caption{Picnic plating task demonstration. The quadruped robot performs a lean-down motion and executes damage-free grasping of a kiwi, followed by organizing and arranging the items. }
  \label{fig:picnic_demo}
\end{figure}
}

\textbf{Task 3: Delicate Object Interaction.}
Compared to wheeled robots, quadruped
robots have better mobility, such as on uneven ground and grass \cite{xiong2020autonomous, droukas2023survey}.
% There have been a number of studies on fruit harvesting with mobile robots (e.g., strawberry picking). However, due to the limitations of wheeled mobile bases, a quadruped robot can move more flexibly over uneven and low-friction terrain such as grass. 
For bruise-prone fruits, end-effector grasping must maintain an appropriate contact force to avoid damage. Beyond testing with a variety of fruits, including bruise, e.g. prone kiwifruit and nectarines, we further demonstrate our capability with a more extreme example: grasping fragile potato chips that can easily crumble. In outdoor environments, we showcase fruit grasping and harvesting capabilities (Fig.~\ref{fig:picnic_demo}). We further conduct controlled lab evaluations and compare against a tactile-free baseline on delicate handling tasks with nectarines, kiwi, and potato chips.  Due to the lack of specialized equipment for detecting fruit bruising, we report quantitative results only for chip manipulation and present the fruit tasks qualitatively. 

\subsubsection{Baseline Design} 
We design these baselines to answer three questions: (a) how much tactile feedback contributes to performance? 
(b) Is our tactile-aware diffusion policy (DP) more effective than a vanilla visuotactile DP?
and 
(c) how much the tactile-aware low-level policy contributes to overall performance?

As shown in Table~\ref{tab:policy_compare}, we design following methods to answer the above questions:

\textbf{Visuomotor Policy (P1).} The entire system does not use tactile feedback. The high-level policy is conditioned only on vision and proprioception and outputs position-only command sequences, which are tracked by a position-tracking low-level controller, same as UMI on legs~\cite{ha2024umi}. 

\textbf{Vanilla Visuotactile DP + Vanilla Low-Level Policy (P2).} We additionally provide tactile observations as conditioning inputs to the high-level policy, but the high-level policy still outputs position-only command sequences. The low-level controller remains position-tracking.

\textbf{Tactile-Aware DP + Vanilla Low-Level Policy  (P3).} The high-level policy is a tactile-aware diffusion policy that outputs both position commands and tactile command sequences. However, the low-level controller does not use the tactile commands and tracks only the position commands. 

\textbf{Tactile-Aware DP + Tactile-Aware Low-Level Policy (Ours, P4).} Our full method uses a tactile-aware diffusion policy at the high level to output position and tactile command sequences, and a low-level policy that takes both position- and tactile-based commands as inputs and tracks them jointly Table~\ref{tab:policy_compare}.

\subsection{Experimental Results and Analyze}

\begin{table}[t]
\centering
\setlength{\tabcolsep}{3pt}
\renewcommand{\arraystretch}{1.15}
\caption{Comparison of \textbf{P4 (ours)} and baseline variants P1--P3 with respect to the components included in the high-level and low-level policies.}
\begin{tabular}{c|ccc|cc}
\hline
\multirow{2}{*}{Policy} & \multicolumn{3}{c|}{High-Level Policy} & \multicolumn{2}{c}{Low-Level Policy} \\
\cline{2-6}
& \begin{tabular}[c]{@{}c@{}}Visuomotor\\Observation\end{tabular}
& \begin{tabular}[c]{@{}c@{}}Tactile\\Observation\end{tabular}
& \begin{tabular}[c]{@{}c@{}}Tactile\\Prediction\end{tabular}
& Vanilla
& \begin{tabular}[c]{@{}c@{}}Tactile-\\Aware\end{tabular} \\
\hline
P1 & \textcolor{green!60!black}{\cmark} & \textcolor{red}{\xmark} & \textcolor{red}{\xmark} & \textcolor{green!60!black}{\cmark} & \textcolor{red}{\xmark} \\
P2 & \textcolor{green!60!black}{\cmark} & \textcolor{green!60!black}{\cmark} & \textcolor{red}{\xmark} & \textcolor{green!60!black}{\cmark} & \textcolor{red}{\xmark} \\
P3 & \textcolor{green!60!black}{\cmark} & \textcolor{green!60!black}{\cmark} & \textcolor{green!60!black}{\cmark} & \textcolor{green!60!black}{\cmark} & \textcolor{red}{\xmark} \\
\rowcolor{gray!15}
\textbf{P4} & \textcolor{green!60!black}{\cmark} & \textcolor{green!60!black}{\cmark} & \textcolor{green!60!black}{\cmark} & \textcolor{red}{\xmark} & \textcolor{green!60!black}{\cmark} \\
\hline
\end{tabular}
\label{tab:policy_compare}
\end{table}

\begin{table}[t]
\centering
\setlength{\tabcolsep}{4pt}
\renewcommand{\arraystretch}{1.15}
\caption{Success rates on 20 rollouts. Task~1 is long-horizon, and we evaluate re-orientation, insertion, and the full sequence independently under identical initial conditions across methods.}
\begin{tabular}{lccccc}
\hline
 & \makecell{Task 1\\Re-orientation} & \makecell{Task 1\\Insertion} & \makecell{Task 1\\Whole} & Task 2 & Task 3 \\
\hline
P1   & 0.35 & 0.15 & 0.05 & 0.35 & 0.20 \\
P2   & 0.95 & 0.70 & 0.55 & 0.80 & 1.00 \\
P3   & 1.00 & 0.70 & 0.60 & 0.80 & 1.00 \\
\rowcolor{gray!15}
\textbf{P4} & \textbf{1.00} & \textbf{0.85} & \textbf{0.80} & \textbf{0.85} & \textbf{1.00} \\
\hline
\end{tabular}
\label{tab:results}
\end{table}

\begin{figure*}[ht]
  \centering
  \includegraphics[width=0.925\textwidth]
  {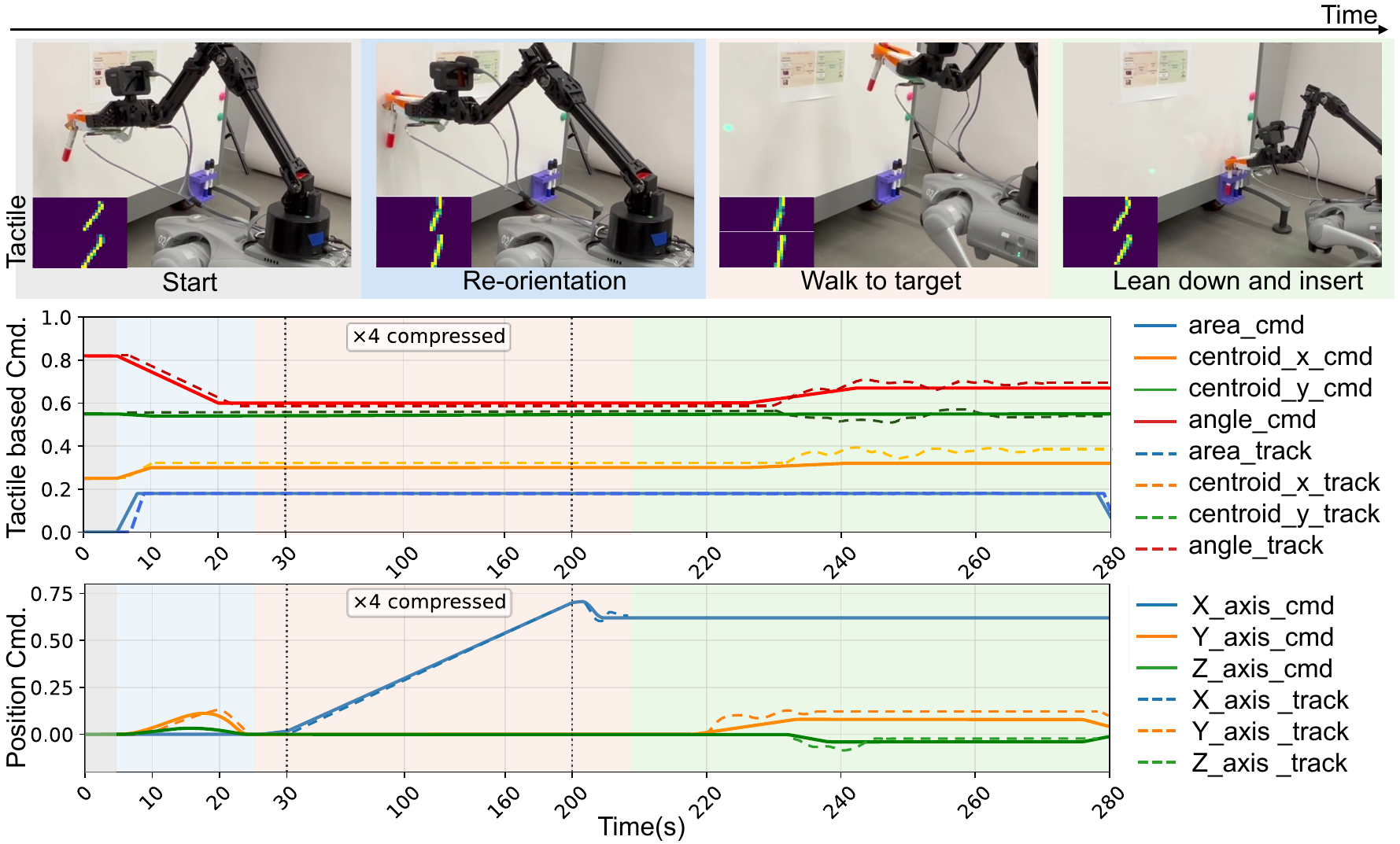}
  \caption{Representative successful autonomous rollout for Task 1 (Extrinsic-Contact-Based Reorientation and Insertion). The robot starts from the reorientation stage (\textcolor{stageBlue}{light blue background}), then moves toward the target socket (\textcolor{stageRed}{light red background}), performs a lean-down motion and completes the insertion task (\textcolor{stageGreen}{light green background}).  The middle time interval is compressed by $4\times$. }
  \label{fig:task1}
\end{figure*}

\begin{figure*}[t]
  \centering
  \includegraphics[width=0.925\textwidth]
  {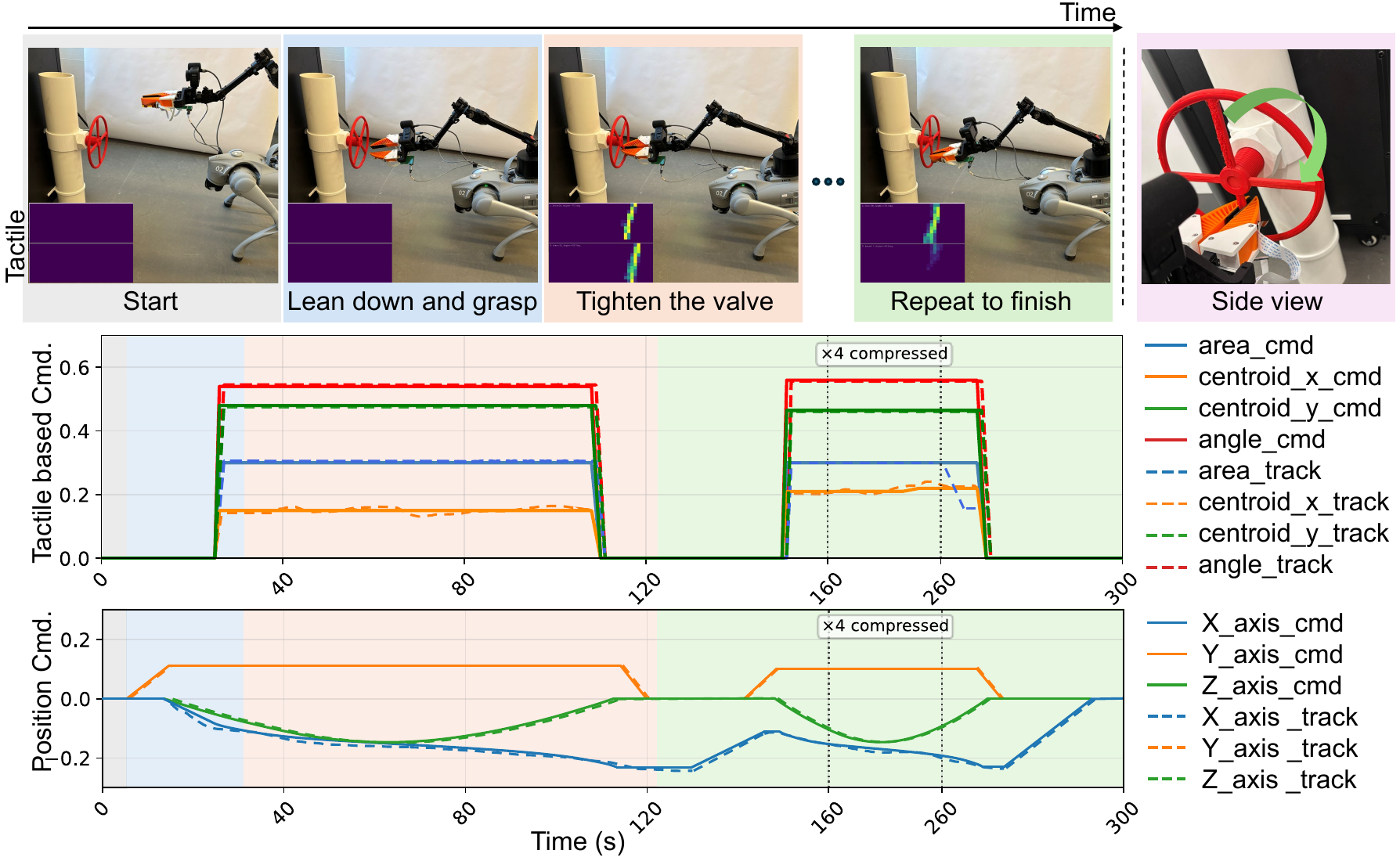}
  \caption{Representative successful autonomous rollout for Task~2 (Valve Tightening). The robot leans down and grasps the valve (\textcolor{stageBlue}{light-blue background}), maintains a near-vertical grasp to tighten (\textcolor{stageRed}{light-red background}), and repeats the tightening cycle until completion (\textcolor{stageGreen}{light-green background}). }
  \label{fig:task2}
\end{figure*}

\begin{figure*}[t]
  \centering
  \includegraphics[width=0.98\textwidth]
  {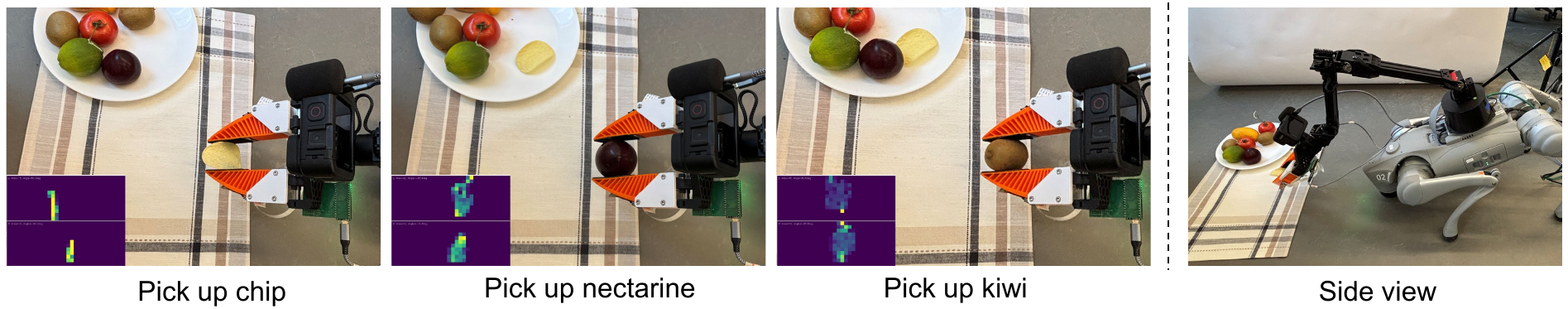}
  \caption{Demonstration of Task 3 (Delicate Object Interaction). The learned policies successfully manipulate a variety of fruits and chips, requiring careful control of grasp force.}
  \label{fig:task3_2}
\end{figure*}

Fig.~\ref{fig:task1} and Fig.~\ref{fig:task2} show the commanded (Cmd.) and tracked (Track) trajectories, where Track denotes the measured value; position on the y-axis is in meters and defined as displacement relative to the initial pose. For visualization, we normalize all plotted signals to the range $[0,1]$; tactile signals are normalized using min--max scaling. For position tracking, the UMI gripper trajectory is defined in terms of relative motion. We use the UMI gripper’s initial pose as the reference (initial) position. During locomotion, we estimate the traveled distance based on the velocity measured by the quadruped’s IMU sensor. In all other cases, we compute the end-effector position using the arm base as the origin. As a result, the displacement induced by base tilting is not reflected in this position measurement chart.

As shown in Fig.~\ref{fig:task1}, the policy tracks the commanded motion accurately and makes adjustments to successfully complete the task, as illustrated in the insertion stage.
From the results in Table~\ref{tab:results}, introducing tactile feedback leads to substantial performance gains over the no-tactile baseline.  Moreover, by comparing Baseline~2 (P2) and Baseline~3 (P3), we verify that using a tactile-aware diffusion policy at the high level does not degrade the position-based command sequence: both methods achieve comparable performance. Finally, comparing Baseline~3 (P3) with our full method demonstrates the benefit of incorporating tactile commands into the low-level policy: our method further increases the success rate from 0.70 to 0.85 on Task~1 insertion (+0.15), from 0.60 to 0.80 on Task~1 whole (+0.20), and from 0.80 to 0.85 on Task~2 (+0.05), and achieves 1.00 on Task~3 (compared to 0.20 for P1).

 In Task 1, during re-orientation, the GoPro camera viewpoint cannot reliably capture the marker’s orientation, leading the tactile-free policy to a low success rate (0.35). A typical failure is incomplete re-orientation, where the marker is not brought to an approximately vertical pose relative to the gripper for subsequent insertion. During insertion, without a low-level controller that tracks tactile-based commands, online adjustment of the marker pose and position becomes difficult, leading to incomplete insertion. In contrast, our method continuously adjusts the pose and successfully completes the task.

In Task 2, policies without tactile feedback cannot reliably infer from vision alone whether the valve is fully tightened, leading to more failures. Valve tightening also requires maintaining vertical contact while continuously adjusting the gripper pose during rotation. Baseline 3 (P3), whose low-level controller does not track tactile-based commands, performs slightly worse than our approach (5\% lower success rate).

Fig.~\ref{fig:task3_2} illustrates our Task~3 evaluation on common food items, including bruise-prone nectarines and kiwifruit, as well as highly fragile potato chips. Since we lack specialized equipment to reliably measure fruit bruising, we report quantitative success rates only for the chip interaction task (as shown in Table~\ref{tab:results}. With tactile feedback enabled, our method achieves a 100\% success rate on potato chip handling. Applying inappropriate contact forces can easily fracture or crumble the chips, highlighting the necessity of tactile feedback for damage-free interaction.

%% file: sections/5conclusion.tex
\section{Discussion and Limitation}
\label{sec: conclusion}

% In this paper, we present a tactile-aware hierarchical loco-manipulation framework that learns from human demonstrations and integrates tactile sensing into both high-level
% planning and low-level execution on a quadrupedal mobile
% manipulator. Overall, our method achieves a 0.2854 average
% improvement across the evaluated contact-rich tasks.

Our results demonstrate that tactile-aware hierarchical learning can improve quadrupedal loco-manipulation. Nevertheless, our current framework still has several limitations. First, tactile observations are not yet encoded in a fully integrated manner, and the current representation may not capture the most compact and informative structure of raw tactile signals. A promising direction for future work is therefore to explore integrated tactile representation learning, for example by compressing raw tactile signals into low-dimensional targets with a tactile variational autoencoder (VAE) encoder to better distill task-relevant information~\cite{van2016stable}.

Second, the current low-level control framework still has limited capability for fine-grained, precise contact-rich loco-manipulation, especially when both whole-body balance and manipulation accuracy must be maintained under environmental uncertainty. Future work should develop a stronger low-level framework that relaxes constraints on the quadruped body. Such a framework could further expand the robot’s effective workspace and enable contact-rich manipulation in more challenging environments, such as sand or soft grass, where the quadruped must continuously adjust its footholds to remain balanced while still maintaining a stable end-effector for precise manipulation.